\pdfoutput=1
\documentclass[final]{cvpr}

\usepackage{times}
\usepackage{epsfig}
\usepackage{graphicx}
\usepackage{amsmath}
\usepackage{amssymb}
\usepackage{nicefrac}
\usepackage{multirow}
\usepackage{xcolor}
\usepackage{pifont}
\usepackage{makecell}
\usepackage{bbm}
\usepackage{arydshln}

\pdfoutput=1
\usepackage{xpunctuate}

\usepackage{xcolor-material}
\usepackage{color, colortbl}

\usepackage{soul} 
\usepackage{xspace}
\usepackage{amsmath}

\usepackage{hyphenat}

\makeatletter
\def\thanks#1{\protected@xdef\@thanks{\@thanks
        \protect\footnotetext{#1}}}
\makeatother

\DeclareRobustCommand{\methodFull}
{Cluster-driven Graph Federated Learning } 

\DeclareRobustCommand{\methodShort}
{FedCG}

\newcommand{\real}{{\rm I\!R}}

\usepackage[pagebackref=true,breaklinks=true,colorlinks,bookmarks=false]{hyperref}

\def\cvprPaperID{67} %
\def\confYear{Learning from Limited and Imperfect Data (L2ID)}

\makeatletter
\let\origparagraph\paragraph
\renewcommand\paragraph{\@ifstar{\starparagraph}{\nostarparagraph}}
\newcommand\nostarparagraph[1]
{\origparagraph{#1}\paragraphpostlude}
\newcommand\starparagraph[1]
{\paragraphprelude\origparagraph*{#1}\paragraphpostlude}
\newcommand\paragraphprelude{%
  \vspace{-14pt}%
}
\newcommand\paragraphpostlude{%
}
\begin{document}

\title{\vspace{-15pt}Cluster-driven Graph Federated Learning over Multiple Domains}

\author{
Debora Caldarola\textsuperscript{*,1}, \quad
\thanks{\textsuperscript{*}Corresponding author: \tt debora.caldarola@polito.it}
Massimiliano Mancini\textsuperscript{2}, \quad
Fabio Galasso\textsuperscript{3}, \quad \\
Marco Ciccone\textsuperscript{4}, \quad
Emanuele Rodolà\textsuperscript{3}, \quad
Barbara Caputo\textsuperscript{1,5} \\

\tt \small \textsuperscript{1} Politecnico di Torino, \textsuperscript{2} University of Tübingen, \textsuperscript{3} Sapienza University of Rome, \\
\tt \small \textsuperscript{4} Politecnico di Milano, \textsuperscript{5} Italian Institute of Technology
}

\maketitle

\pagenumbering{gobble}

\begin{abstract}
Federated Learning (FL) deals with learning a central model (\ie the server) in privacy-constrained scenarios, where data are stored on multiple devices (\ie the clients). The central model has no direct access to the data, but only to the updates of the parameters computed locally by each client. This raises a problem, known as \textit{statistical heterogeneity}, because the clients may have different data distributions (\ie domains). This is only partly alleviated by clustering the clients. Clustering may reduce heterogeneity by identifying the domains, but it deprives each cluster model of the data and supervision of others.

Here we propose a novel \methodFull (\methodShort). In \methodShort, clustering serves to address statistical heterogeneity, while Graph Convolutional Networks (GCNs) enable sharing knowledge across them. 
\methodShort:\ \textbf{i.}\ identifies the domains via an FL-compliant clustering and instantiates domain-specific modules (residual branches) for each domain;
\textbf{ii.}\ connects the domain-specific modules through a GCN at training to learn the interactions among domains and share knowledge;
and \textbf{iii.}\ learns to cluster unsupervised via teacher-student classifier-training iterations and to address novel unseen test domains via their domain soft-assignment scores.
Thanks to the unique interplay of GCN over clusters, \methodShort\ achieves the state-of-the-art on multiple FL benchmarks.
\end{abstract}

\pdfoutput=1
\vspace{-15pt}
\section{Introduction}

\begin{figure}[t]
\begin{center}
   \includegraphics[width=1\linewidth]{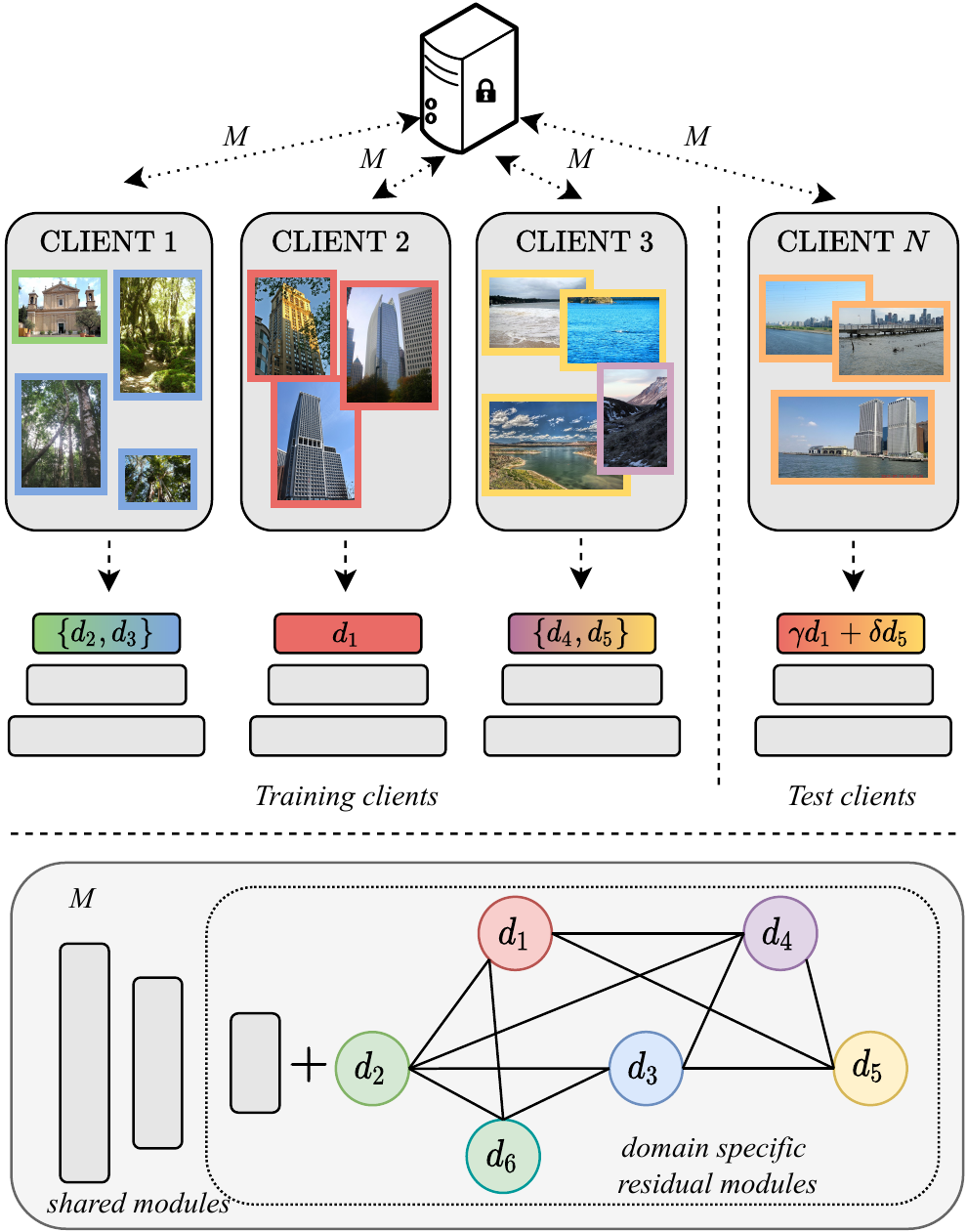}
\end{center}
  \vspace{-5pt}
   \caption{In a federated scenario, clients and server exchange the parameters of the model $M$. Each client has access to its local data, which can be non-i.i.d. and unbalanced. In the image, each color identifies a different distribution, \ie a domain, such as pictures of skyscrapers or sea landscapes. Our model $M$ is made of domain-agnostic layers (in gray) and a GCN containing domain-specific parameters, added as residual. According to the domains of the input images, the corresponding nodes of the GCN are activated. At test time, new domains can be addressed as a soft combination of the discovered ones, \eg skyscrapers over the sea.}
 \vspace{-5pt}
\label{fig:teaser}
\end{figure}

In Federated Learning (FL)~\cite{mcmahan2017communication}, a central server model is trained using data stored locally on multiple client devices. Each client computes a local update of the model and all the client updates are then aggregated server-side to build the final model. %
Since no data ever leaves the client devices, %
the central model has no direct access to the raw data itself, a fundamental requirement for privacy-preserving applications (e.g.\  medical records, bank transactions, etc.).  %
FL usually relies upon the key assumption that a single central model can work efficiently across several users ~\cite{kairouz2019advances,li2020federated}. This may not hold in practice, since distinct clients might hold different input distributions, \ie \textit{domains} (\eg, people speaking different languages, pictures taken at different locations), with their data possibly being not identically distributed and/or unbalanced. 
These issues (collectively referred to as \textit{statistical heterogeneity}~\cite{smith2017federated}) imply that all i.i.d. assumptions made in distributed optimization or centralized training are violated, and modeling the learning problem requires taking into account this new complexity.

To the best of our knowledge, statistical heterogeneity has been tackled so far with diverse approaches, but none of them has modeled the direct share of knowledge between domains. In particular, meta-learning FL techniques focused on the client-server relation~\cite{jiang2019improving,NEURIPS2019_f4aa0dd9,NEURIPS2020_24389bfe,chen2018federated}; multi-task FL methods specialized parts of the models to certain clients~\cite{smith2017federated,corinzia2019variational}; while 
clustering-based FL split the clients and data, learning separate models for them~\cite{sattler2020clustered,xie2020multi}.

In this work, we introduce a novel \methodFull (\methodShort). \methodShort\ leverages clustering and its potential to reduce statistical heterogeneity by identifying homogeneous\footnote{Homogeneous stands in this context for groupings that minimize intra-cluster Vs inter-cluster variance.}. Concurrently, \methodShort\ is the first to model the domain-domain interaction by means of a GCN, which connects domain-specific model components. In the GCN, each node consists of domain-specific model parameters, while the adjacency matrix is composed of the inverse pairwise distances between the domain-specific parameters.
In this way, \methodShort\ not only captures the specificity of each domain but also allows each domain to benefit from the updates of others, sharing knowledge at training.

Our clustering is based on unsupervised teacher-student~\cite{hinton2015distilling} classifier-training iterations and it generalizes to unseen test-time domains. We cluster by pseudo-labels, assigned by a teacher and learned by a student, in rounds of refinements. This is accomplished within the FL training paradigm, respecting the client's privacy. This allows to estimate soft-assignments for unseen novel test domains.

We test our model extensively on several FL benchmarks, demonstrating results above or competitive with the state-of-the-art. %
Our main contributions are: %
\begin{enumerate}
    \item We present the first cluster-driven GCN-based approach to address statistical heterogeneity in the FL scenario. Thanks to the interactions among domains learned by the means of a GCN, knowledge is shared across domains according to a similarity-based criterion, reducing the risk of overfitting and helping the less populated domains.
    \item We introduce an iterative teacher-student clustering algorithm designed for the federated learning scenario, which allows adapting to new domains via soft-assignments. This captures the diverse domain distributions without violating the FL constraints. Each domain is assigned model-specific components, trained via GCN interactions.
    \item We evaluate our model on multiple FL benchmarks, where we compare favorably or on par with respect to the state-of-the-art.
\end{enumerate}
\pdfoutput=1
\section{Related Work}
\vspace{-5pt}
Although FL is a relatively new field of study, it has aroused great interest in the research community because of its wide applicability in privacy-constrained scenarios~\cite{li2020federated}.
A simple but effective baseline for FL is the Federated Averaging (FedAvg) algorithm~\cite{mcmahan2017communication}, %
where the central model is obtained as %
a weighted average of the models received from each client after their local updates. %
FedAvg has been extensively studied and extended by changing either what is averaged, or how the local models are considered in each update. For instance, FedSGD~\cite{mcmahan2017communication} bases the update on the model's gradient instead of the weights, while in~\cite{konevcny2016federated}, the updates are parametrized with fewer variables to reduce the uplink communication cost. %
Similarly, in~\cite{reisizadeh2020fedpaq} %
the nodes only send a quantized version of their local information for reducing the communication overhead. Mohri \etal~\cite{pmlr-v97-mohri19a} propose Agnostic Federated Learning (AFL) in order to reduce the bias towards specific clients. In~\cite{li2019fedmd}, each client designs its own local model and the local information is shared by means of knowledge distillation. %
Our work mainly relates to \cite{mcmahan2017communication}, but we explicitly revise the FedAvg framework to account for statistical heterogeneity. %

\paragraph*{Statistical heterogeneity in FL} 
Despite their effectiveness, the previous methods ignore an important problem of FL, \ie statistical heterogeneity.
Many works~\cite{li2018federated,hsu2019measuring,li2019feddane,karimireddy2020scaffold} study this challenge in terms of convergence analysis and effects of non-i.i.d. data distributions in the federated scenario.
Others address this problem from the meta-learning~\cite{nichol2018first} and multitask~\cite{caruana1997multitask} perspectives for building specialized models. 
Specifically, {Model-Agnostic Meta-Learning}~\cite{finn2017model} caught the interest of the FL community for its compatibility with any ML model trained with gradient descent~\cite{jiang2019improving,NEURIPS2019_f4aa0dd9,NEURIPS2020_24389bfe,chen2018federated}.
On the other hand, in {Federated Multi-Task Learning} (FMTL)~\cite{smith2017federated,corinzia2019variational}, each client is seen as a different task. FMTL addresses underlying similarities and structures common to some clients %
by learning a separate model for each device of the network. In particular,~\cite{sattler2020clustered} and~\cite{xie2020multi} cluster clients according to their data distribution, assigning a specialized model to each cluster: \cite{sattler2020clustered} uses the cosine similarity of the clients' gradient update, while~\cite{xie2020multi} dynamically groups clients exploiting structural similarities. Hsu \etal~\cite{hsu2020federated} develop FedIR and FedVC for re-sampling and re-weighting the client pools. %

As in the aforementioned algorithms, in this work, we focus on %
addressing statistical heterogeneity in FL. Similarly to~\cite{sattler2020clustered,xie2020multi}, %
we seek to learn specialized models addressing the different data distributions. %
However, differently from~\cite{sattler2020clustered}, our clusters are built and updated {\em during} the federated communication rounds through a domain classifier, and not offline after the model convergence. This removes the requirement that all client-specific models must be stored server-side. %
Moreover, our method aims to identify and group distributions rather than the clients themselves, allowing us to model the realistic case where a client's data may belong to multiple distributions. %
In addition, while the applicability of~\cite{xie2020multi} is constrained to the clients seen during training, our domain model can be applied to unseen clients at test time thanks to the flexibility of our domain classifier. Finally, we explore merging FL with Graph Representation Learning to address statistical heterogeneity, using graphs to learn domain-specific parameters and to model the interactions among them.

\paragraph*{Graph Representation Learning} 
Our work employs Graph Convolutional Networks~\cite{kipf2016semi}, using domain-specific parameters as high-dimensional features at each node of the graph. This is partly inspired  from~\cite{mancini2019adagraph}, where domain-specific batch normalization~\cite{ioffe2015batch} layers are connected through a graph for addressing predictive domain adaptation. However, here we focus on a completely different problem (\ie FL), requiring a different training paradigm, and we build our graph on arbitrary layers of the network.

In the context of FL, to the best of our knowledge, the only existing works adopting graphs as auxiliary representations are SGNN~\cite{mei2019sgnn}, ASFGNN~\cite{zheng2021asfgnn} and GraphFL~\cite{wang2020graphfl}. The first two employ graphs for a different purpose: they use a similarity-based graph neural network for improving node classification in network embeddings, while preserving user privacy. GraphFL, instead, is a semi-supervised node classification method on graphs and uses the FL scenario to solve real-world graph-based problems.  {Differently from these works, we use the graph-based formulation not to learn a %
single general model, but to %
capture the statistical heterogeneity while taking into account the relations among the different data distributions. Thanks to the %
graph, each domain can be addressed with specific parameters while still taking advantage of all local updates.}%

\pdfoutput=1
\section{\methodFull}
In this section we present our approach addressing statistical heterogeneity in FL {by means of GCNs}. Our method is based on three ideas: i) identify the clusters of data sharing the same distribution, ii) assign specific network components to each cluster, and { iii) let the components interact within a GCN}. We name our full model \methodFull (\methodShort). Before describing \methodShort, in the following we formalize the FL problem.

\paragraph*{Problem Formulation.}
\label{sec:problem_formulation}
Our goal is to learn a function $f_\theta:\mathcal{X}\rightarrow\mathcal{Y}$, parametrized by $\theta$, mapping samples from an input space $\mathcal{X}$ to their corresponding semantic in an output space $\mathcal{Y}$. Specifically, we focus on a classification task, where $\mathcal{X}$ contains images while $\mathcal{Y}$ is a probability simplex defined over a set of labels $Y$. %

In the FL setting, the server does not have direct access to the data, but can communicate with a set $C$ of clients, where each client $c\in C$ has access to a local dataset $\mathcal{T}_c=\{x_i,y_i\}_{i=1}^{n_c}$ with $x\in \mathcal{X}$ and $y\in {Y}$. In this scenario, we can learn $f_\theta$ by querying clients and relying on their local updates of the parameters $\theta$. In particular, since $|C|$ is large, we can assume a synchronous update scheme proceeding in communication rounds, where in each round a set $K$ of clients receives $f_\theta$, with $|K|\ll |C|$. Each client $k\in K$ computes a local update of $\theta$, \ie $\theta_k$, with its local dataset $\mathcal{T}_k$, by minimizing a given objective function. Since we consider classification tasks, we update $\theta_k$ by minimizing the standard cross-entropy loss over $\mathcal{T}_k$:
\begin{equation}
    \label{eq:client-updates}
    \theta_k = \min_\theta -\frac{1}{n_k}\sum_{(x,y)\in \mathcal{T}_k} \log{f^y_\theta(x)} \,,
\end{equation}
where $f^y_\theta(x)$ denotes the probability of $x$ to belong to class $y$ as given by $f_\theta$. 

With Eq.~\eqref{eq:client-updates}, we obtain for each client $k\in K$ its corresponding local parameters $\theta_k$ tuned to address the classification task of $\mathcal{T}_k$. At each round, the server gathers all local updates %
and combines them to update the central model parameters $\theta$. A simple yet effective strategy to aggregate the local updates is FedAvg~\cite{mcmahan2017communication}, that computes $\theta$ as the weighted average of each $\theta^k$:
\begin{equation}
    \label{eq:fedavg}
    \theta = \frac{1}{\sum_{k\in K} n_k} \sum_{k\in K} n_k \theta_k \,.
\end{equation} %

Heterogeneity may be a problem in FedAvg, and in general for FL strategies, due to the lack of convergence guarantees in non-i.i.d. and unbalanced data~\cite{smith2017federated,li2020federated}. %
In realistic applications, the joint probability distributions over $\mathcal{X}$ and ${Y}$ are usually different in each client, \ie given two clients $c$ and $k$ with $c\neq k$, we have $p_{\mathcal{X}Y}(\mathcal{T}_c)\neq p_{\mathcal{X}Y}(\mathcal{T}_k)$. 

To address this problem, 
we propose an approach that i) identifies the distributions (\ie domains) present in different clients through clustering; ii) instantiates domain-specific components to adapt the model to each domain; iii) makes the various domain-specific modules interact through a GCN, such that updating one of them can benefit the others. In the following we describe each of these elements.

\subsection{Federated Clustering}
\label{sec:clustering}
To address statistical heterogeneity through domain-specific modules, we need to identify the different domains present in the data. This is challenging since %
data are split across multiple clients and the server cannot cluster them directly. %
Moreover, %
these clusters, even if correctly identified for the training set, may not be optimal for the test set.
Here we address the first problem by a clustering procedure built on two \textit{domain classifiers}, one having the role of the \textit{teacher} and the other of the \textit{student}, which 
iteratively group images such that their grouping is easier to classify.
We describe how we match clusters to the test set in Sec.~\ref{sec:method-graph}. %

Formally, let us assume our data contain $D$ domains, with $D$ being a hyperparameter.  We initialize two domain classifiers, the teacher $g_\phi$ and the student $g_\varphi$ parametrized by $\phi$ and $\varphi$ respectively. Each domain-classifier is a function, mapping images to a probability vector $\mathcal{D}$ defined over the $D$ domains, \ie $g_\cdot: \mathcal{X} \rightarrow \mathcal{D}$. %
Given an input image, the teacher provides domain pseudo-labels as a target to refine student's predictions. In particular, we learn the client student parameters $\varphi_k$ by iteratively  minimizing the cross-entropy loss between the teacher and student domain predictions over $\mathcal{T}_k$.  Thus, for a client $k\in K$, the parameters $\varphi_k$ of the student are:
\begin{equation}
    \label{eq:client-updates-domain-classifier}
    \varphi_k = \arg\min_\varphi -\frac{1}{n_k}\sum_{(x,y)\in \mathcal{T}_k}\log{g^{\hat{d}}_\varphi(x)} \,,
\end{equation}
where $\hat{d}$ is the pseudo-label given by the teacher for $x$, \ie $\hat{d} = \arg\max_{d\in D} g_\phi^d(x)$ and $g_{*}^d(x)$ denotes the probability of $x$ to belong to the $d$-th domain as given by $g_{*}$.  Eq.~\eqref{eq:client-updates-domain-classifier} rewards the student from being able to classify according to the pseudo labels, and implicitly encourages agreement on the pseudo-labels, thus on the clustering, which most easily may be agreed upon.
Then the domain classifier parameters $\varphi$ are updated after each round with standard FedAvg, \ie $\varphi = \frac{1}{\sum_{i=1}^K n_k} \sum_{k\in K} n_k \varphi_k$.

The idea behind this approach is inspired from deep clustering with self-labelling~\cite{YM.2020Self-labelling}, i.e.\ the teacher and the student networks would find the equilibrium once they group images in such a way so they can be more easily recognized. This reconnects to the DNNs being natural deep image priors\cite{UlyanovCVPR18}, working well for image-related tasks even if just randomly initialized. And it may intuitively match that a ``bad'' labelling would leave no alternative to a DNN but to overfit~\cite{ZhangICR17gen}, which may be hard to imitate by the student.
Differently from~\cite{YM.2020Self-labelling}, since we have no access to data and cluster labels, we use the teacher $g_\phi$ to provide them locally in each client. Both $\phi$ and $\varphi$ are randomly initialized and $\phi$ is fixed during training. After $T$ rounds, with $T$ being a hyperparameter, the parameters $\phi$ of the teacher are updated with the current student ones $\varphi$, iteratively. 
 
 Note that, unlike previous works~\cite{xie2020multi}, our clustering algorithm can assign unseen data to clusters at test time, thanks to the domain classifier. In particular, the cluster assignment of a test image $x$ corresponds to the domain probabilities given by the student $g_\varphi$. Since $g_\varphi(x)$ is soft, we can accommodate for data belonging to unseen domains by a combination of existing ones. %
 Additionally, in our formulation, one client's data samples may belong to multiple clusters, considering the more general case where each client may contain more than one data distribution. %

\subsection{Cluster-specific Models}
\label{sec:specific-models}

Since our model can identify data clusters through the previously described procedure, we can design a way to specialize the function $f_\theta$ to each domain. Inspired by multi-domain learning~\cite{rebuffi2017learning,rosenfeld2018incremental,rebuffi2018efficient,mallya2018piggyback,mancini2020boosting}, we can achieve this with domain-specific components. For simplicity, let us consider the parameters $\theta$ to be split into two sets, \ie $\theta = \{\theta_a, \theta_s\}$ where $\theta_a$ are the domain-agnostic parameters and $\theta_s$ the domain-specific ones. Note that $\theta_s$ is actually a set $\theta_s=\{\theta^d_s\}_{d=1}^D$ where $\theta_s^d$ are the parameters specific to the $d$-th domain. To tailor the model to a specific domain, we can consider multiple ways to include $\theta_s$, such as direct influence on the agnostic parameters $\theta_a$~\cite{rosenfeld2018incremental,mallya2018piggyback,mancini2020boosting} or residual activations~\cite{rebuffi2017learning,rebuffi2018efficient}. Here we follow the latter strategy, since the former relies on the robustness of $\theta_a$, which is harder to guarantee in FL.
Let us assume $f_\theta$ to be a deep neural network with a set of layers $L$, denoting as $f^{\ell}_{\theta}$ the function applied at layer $\ell\in L$. Given input from a domain and the features $z^{\ell}$ extracted at the previous layer, the output of the $\ell$-th layer is:
\begin{equation}
    \label{eq:domain-specific-layers}
    z^{\ell} = f^{\ell}_{\theta_a}(z) + \lambda_l \sum_{d=1}^{D}w_d\cdot f^{\ell}_{\theta_s^d}(z) \,,
\end{equation}
where $\lambda_l$ is a learnable parameter balancing the effect of the domain-specific components and $w_d$ is the weight of domain $d$. During training, we assume data to belong to a single cluster, given by the pseudo-labels of the teacher, thus $w_d$ is 1 if $d=\hat{d}$ and 0 otherwise. At test time, we want our model to deal with data from arbitrary domains by simply combining residuals of seen ones. Thus we set $w_d=g^d_\varphi(x)$, weighting the impact of each domain-specific component by the student output probabilities. %
Note that the formulation in Eq.~\eqref{eq:domain-specific-layers} is general, with $f^{\ell}_{\theta}$ being any layer of a standard convolutional neural network. We explored its application to either the whole network or the last layers.

Since we are in a federated scenario, also the central domain-specific parameters must be updated without access to local data and after each round. In practice, we follow Eq.~\eqref{eq:fedavg} and we perform FedAvg on both domain-agnostic and domain-specific parameters in each training round.

\begin{figure*}
\begin{center}
    \includegraphics[width=\linewidth]{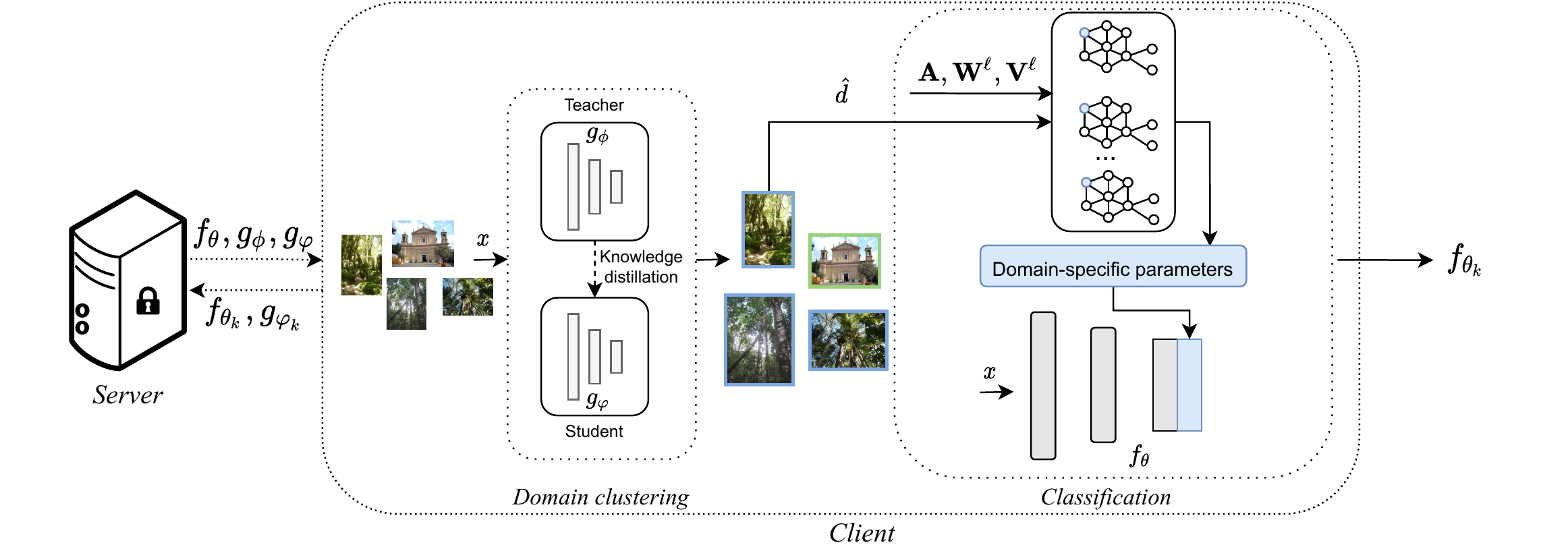}
\end{center}
\vspace{-12pt}
   \caption{\methodShort\ framework (best seen in colors). The server sends the model $f_\theta$ to the clients selected for the federated round, together with the teacher $g_\phi$ and student $g_\varphi$ domain classifiers. On the client-side, the domain classifier clusters the local data $x$, producing as output the domain of belonging $\hat{d}$ of each image. At training time, the hard label $\hat{d}$ is predicted by $g_\phi$ and is used as input to train $g_\varphi$ through a process based on knowledge distillation. At test time, $\hat{d}$ is given by $g_\varphi$ and is a weighted combination of the discovered domains. In \methodShort, the network $f_\theta$ is made of a domain-agnostic part (in gray) and a residual domain-specific one (in blue). The domain-specific parameters are produced by the GCN, receiving as input  $\textbf{A},\textbf{W}^\ell, \textbf{V}^\ell$ and $\hat{d}$. After training both $f_\theta$ and $g_\varphi$ on its data, the client $k$ sends back to the server the updated weights $\theta_k$ and $\varphi_k$. On the server-side, the updates are aggregated by the means of the FedAvg algorithm.}
\label{fig:model}
\end{figure*}

\subsection{Connecting Cluster-specific Models}
\label{sec:method-graph}
We now have a model that can adapt to the specificity of each domain. Here we propose to refine the domain-specific parameters by making them interact.
Specifically, we model the interaction of the domain-specific parameters of each layer $\ell$ via a graph $\mathcal{G}^{\ell}=(\mathcal{V}^{\ell},\mathcal{E}^{\ell})$, where the nodes $i\in\mathcal{V}^{\ell}$ are the set of all domain-specific parameters at layer $\ell$, and $e_{ij}\in\mathcal{E}^{\ell}$ are the edges connecting two domain nodes $i$ and $j$ which may interact together. This addresses the drawback of our formulation in Sec.~\ref{sec:specific-models}, \ie if a domain has few assigned samples, its parameters will be rarely updated and thus not robust enough to capture the specificity of the domain and generalize to unseen samples of the same domain.

{We propose to use a GCN~\cite{kipf2016semi} to model the interaction of domain-specific parameters. }
Let us collect in the matrix $\mathbf{V}^{\ell}$ the value of each node, \ie all domain-specific parameters at layer $\ell$: %
$\mathbf{V}^{\ell}= [\theta^1_{s|l},\dots,\theta^D_{s|l}]^\intercal \in \real^{D\times q}$, with %
$q=|\theta^d_{s|l}|$ the number of parameters per domain.  We compute the graph-version $\hat{\mathbf{V}}^{\ell}$ of the domain-specific parameters $\mathbf{V}^{\ell}$ as:
\begin{equation}
\label{eq:gcn-output}
     \hat{\mathbf{V}}^{\ell} = \sigma({\mathbf{A}}\,\, {\mathbf{V}}^{\ell} \,\, \mathbf{W}^{\ell}) \,,
\end{equation}
 where $\sigma$ is an activation function (\eg ReLU), $\mathbf{A} \in \real^{D\times D}$ is the adjacency matrix defined across the domains, and $\mathbf{W}^{\ell} \in \real^{q\times {q'}}$ %
is a weight {projection} matrix, projecting the domain-specific parameters into dimension $q'$. Here, for simplicity, we set $q=q'$. In \methodShort, we replace the domain-specific parameters of Eq.~\eqref{eq:domain-specific-layers} with the ones computed in Eq.~\eqref{eq:gcn-output}. Similarly to all other parameters of the network, we update $\mathbf{W}$ in each training round through FedAvg. In case $q$ is large, we implement $\mathbf{W}$ as a multi-layer bottleneck (see implementation details). %

The values in the adjacency matrix encode, for each edge, how close two domains are; since we have no priors on the structure of the graph, we model $\mathcal{G}^{\ell}$ as a fully-connected weighted graph. Without direct access to the data server-side, we compute the distance among two domains directly in the (domain-specific) parameter's space. In practice, we define the similarity $h_{i,j}$ among domains $i,j$ as:
\begin{equation}
    \label{eq:domain-similarity}
    h_{i,j} = \frac{1}{\Vert \theta^i_{s} - \theta^j_{s}\Vert_2}\,,
\end{equation}
and the corresponding value $\mathbf{A}_{ij}$ in the adjacency matrix as: 
\[ \mathbf{A}_{ij} = 
\begin{cases}
    \beta \,\,\, & \text{if}\,\,\, i=j\\
    \frac{(1-\beta) \cdot h_{ij}}{\sum_{d=1}^D \mathbbm{1}_{i\neq d} h_{id}} &  \text{otherwise} %
    \label{eq:edge_weight}
\end{cases}
\]
where $\beta$ is a hyperparameter weighing the impact of the self-connection, which we set to $0.5$, and $\mathbbm{1}_{i\neq m}$ is an indicator function being 1 when $i\neq m$ and 0 otherwise. 

In our formulation, each client receives not only the set of parameters $\theta$, but also the adjacency matrix. With this definition, we are forcing the gradient of a domain-specific component to flow to all others through the GCN. Consequently, an update on a domain-specific component will influence all domain-specific parameters, even the ones of the domains not present in the current training round. Moreover, given two domains $i,j$ with $i\neq j$, the influence of $j$ on $i$ in each layer is directly proportional to the adjacency matrix value $\mathbf{A}_{ij}$. This means that the more two sets of domain-specific parameters are close, the higher is their mutual influence. %
Finally, while the GCN is a way to ensure information flow across domains during training, at inference we can just precompute $\hat{\mathbf{V}}^{\ell}$ for each layer, to save memory usage.

\pdfoutput=1
\section{Experiments}
\subsection{Datasets and implementation details}
We evaluate the proposed model on image classification tasks on the LEAF benchmark~\cite{caldas2018leaf}, testing both on the CelebA~\cite{liu2015faceattributes} and Federated Extended MNIST (FEMNIST)~\cite{lecun-mnisthandwrittendigit-2010,cohen2017emnist} datasets. %
Table \ref{tab:datasets} details each setting. 

\paragraph*{CelebA} 
is a widely used dataset containing pictures of faces of several celebrities. We follow the same experimental protocol of ~\cite{caldas2018leaf}, partitioning the dataset by celebrities and ignoring the ones with less than 5 images. The task is binary classification, recognizing whether the depicted person is smiling or not. Following \cite{caldas2018leaf}, we use $10\%$ of the total clients for training and a separate split of $20\%$ of them for test. We train FedAvg and our model on 100 rounds with 10 clients each, %
training locally for a single epoch with a batch size of $5$ and a learning rate of $10^{-3}$. 
To perform a fair comparison we used the same architecture of \cite{caldas2018leaf}, replacing convolutional and batch-normalization layers with their \methodShort\ counterpart, based on a 1-layer GCN. %

\paragraph*{FEMNIST} 
contains images depicting different characters drawn by different writers. The task is a $62$-way classification problem, where the classes correspond to the uppercase and lowercase %
letters of the alphabet and numbers. %
Following the setting proposed by~\cite{caldas2018leaf}, each client corresponds to a different writer, using $60$\% of them for training, $20$\% for validation and $20$\% for test. %
We run both FedAvg and \methodShort\ for $1000$ rounds of $5$ clients each, using a batch size of $10$, a learning rate of $10^{-3}$ and one local epoch. %
We use the same and architecture of \cite{caldas2018leaf}, replacing the last convolutional layer with our GCN-based version. In this case, we use a 2-layers GCN, modeling the projection matrix $\mathbf{W}$ as a bottleneck dividing the features by a factor of $16$.

\paragraph*{Implementation details} In all datasets, the domain classifiers are CNNs made of two convolutional layers of $32$ and $64$ features and kernel size $3\times3$, followed by an average pooling and a linear layer whose output dimension is the number of domains. %
We train the domain classifiers through an SGD optimizer without weight decay and a learning rate of $10^{-4}$. We implement \methodShort\ on PyTorch~\cite{NEURIPS2019_9015}, running the experiments on NVIDIA GeForce 1070 GTX GPUs. We chose PytTorch due to its higher flexibility for prototyping and experimenting the components of our model. To ensure a fair comparison, we implemented the FedAvg baseline using the same framework, architectures, hyperparameters and training protocols of \cite{leaf2018github}.Table~\ref{tab:fedavg} compares our results with the performance of the Tensorflow~\cite{abadi2016tensorflow} implementation from the LEAF repository~\cite{leaf2018github}:
our FedAvg baseline outperforms the original one by almost 3\% in accuracy on FEMNIST, while performing almost 2.5\% less on CelebA. Nevertheless, our main interest is to evaluate the relative improvement of the proposed model with respect to a baseline that does not exploit domain information, using the same aggregation strategy of FedAvg for federated learning. For these reasons, in the following we will take as reference the FedAvg results of the PyTorch framework, to ensure a comparison with the baseline under the exact conditions. %
We evaluate our results in terms of global accuracy on the test set, \ie on the union of the images of all test devices.  %
All experiments on the same dataset were run with the same configuration to perform a fair comparison between the considered approaches.

\begin{table}[t]
    \centering
    \resizebox{\linewidth}{!}{\begin{tabular}{l c c c c c}
        \hline
        \textbf{Dataset} & \textbf{Clients} & \textbf{Total samples} & \multicolumn{2}{c}{\textbf{Samples per client}} & \textbf{Classes}\\
        \cline{4-5}
        & & & Mean & Stdev & \\\hline
        CelebA &  9,343 & 200,288 & 21.44 & 7.63 & 2\\ 
        FEMNIST & 3,550 & 805,263 & 226.83 & 88.94 & 62\\ 
        \hline
    \end{tabular}}
    \vspace{1pt}
    \caption{\textbf{Datasets Statistics}}
    \label{tab:datasets}
\end{table}

\subsection{Ablation study}
\label{sec:ablation}

 In this section, we focus our analysis on testing the performance of the proposed model on the \textit{CelebA} dataset, analyzing the various components of our approach. %
All referred studies and results can be found in Table \ref{tab:ablation_32} and \ref{tab:ablation_cluster}. 

\begin{table}[t]
    
    \centering
    \begin{tabular}{c c c}
        \hline
        \textbf{Dataset} & \textbf{TensorFlow} & \textbf{PyTorch}\\\hline
        CelebA & 89.46 & 86.88 \\
        FEMNIST & 74.72 & 77.81 \\\hline
    \end{tabular}
    \vspace{2pt}
    \caption{Accuracy of FedAvg in TensorFlow \cite{caldas2018leaf} and our version implemented in PyTorch. %
    }
    \vspace{-9pt}
    \label{tab:fedavg}
\end{table}

\vspace{-10pt}
\subsubsection{How to use domain information}
\vspace{-6pt}
To create a sanity-check for our model, we first define the domains manually, exploiting the a priori knowledge given from the images meta-data, \ie the 40 attributes of the dataset (Table \ref{tab:ablation_32}). This allows us to isolate the choice of how to include domain-specific information within the model, without any influence from the clustering procedure. From the 40 attributes, we select the combination of $n$ attributes leading to the most balanced subdivisions of the dataset and %
having a low correlation with the target feature. Since each attribute can only assume the values $\{0,1\}$, the number of possible \textit{domains} is given by all the $2^n$ combinations of the $n$ features. We choose $n=5$, having $N=32$ domains. The selected features are \textit{attractive}, \textit{heavy makeup}, \textit{high cheekbones},  \textit{mouth slightly open} and \textit{wavy hair}. 

\paragraph*{Domain-specific models} We start by replacing the standard single server model with $N$ separate domain-specific models, trained and tested only on the images of their specific domains. As shown in Table \ref{tab:ablation_32}, the performance drops significantly ($33.61$\% vs $86.88$\% of FedAvg%
), since the insufficient amount of data seen by each model leads to poor generalization. This shows that learning a single full model per each domain is not a viable strategy in this scenario.

\paragraph*{Modeling the relations across domains} In order to account for the relations existing among the different domains, we introduce the graph, modeled as a 1-layer GCN. To study the impact of introducing a GCN, we also test a simpler version of the model without the weight transformation matrix $W$. 
We analyze different choices of $\mathbf{A}$, considering the cases where %
i) they are uniformly weighted (U) and ii) the domains are weighted according to a similarity criterion (H), %
specifically the normalized inverse of the Hamming distance~\cite{norouzi2012hamming} between the numerical representation of the domains (\ie their binary metadata).
As Table~\ref{tab:ablation_32} shows, using a GCN consistently improves the final performance over the domain-specific models. The uniform adjacency matrix performs slightly better than the weighted one in this case, with both their performance improving when the projection matrix $\mathbf{W}$ is introduced. These results confirm %
the importance of making the domain-specific nodes interact. However, the results are not satisfactory, being either below or just 1\% above (GCN-H with $\mathbf{W}$) FedAvg. This means that domain information is still dully exploited within the model. %

\paragraph*{Residual domain-specific layers.} %
Finally, we analyze the usage of 
domain-specific parameters to produce residual activations (\ie  Eq.~\eqref{eq:domain-specific-layers}), as in \methodShort, comparing it with the GCN when not using any domain-agnostic component. %
As Table~\ref{tab:ablation_32} shows, while the model with uniform adjacency matrix (U) sees a decrease in performance from GCN to \methodShort\ (\ie 87.92\% vs 86.96\%), the model with weighted adjacency matrix (H) sees a large boost, going from the 84.25\% accuracy of GCN to the 88.65\% of \methodShort. We can draw two conclusions. First, using residual layers to refine the domain agnostic activations (\methodShort) performs better than using only domain-specific components (GCN). Second, when domain-specific components are integrated as residuals, they are much more effective when connected in a weighted (H) rather than a uniform (U) fashion. This is proved by the results of \methodShort-H, surpassing \methodShort-U by 1.7\% in accuracy. %
Finally, we test the importance of the ReLU non-linearity applied to the output of the residual GCN. The non-linearity improves \methodShort, both when the domains are connected uniformly (+1\%) and in a weighted fashion (+0.9\%). The final \methodShort\ model with 1-layer GCN filtered with a ReLU and a weighted adjacency matrix outperforms the baseline FedAvg by 2.6\% accuracy, showing the effectiveness of our choices.

 \vspace{-12pt}
\subsubsection{How to identify the domains}
\vspace{-6pt}
In the previous section, we analyzed how to integrate domain-specific components given oracle domain information. In this section, we drop the assumption of having such information and we %
study the effectiveness of the domains discovered through our clustering procedure on \methodShort\, held out through the %
teacher-student domain classifier
(cf.\ Sec. \ref{sec:clustering}). We report the results of our analysis in Table \ref{tab:ablation_cluster}. 

\paragraph*{Domains extracted through clustering}  %
We start by comparing our domain classifier with the K-means algorithm~\cite{macqueen1967some} applied to the parameters of the models trained separately on each client. %
\methodShort\ performs clustering locally instead, accessing only a subset of the clients at each round. As Table \ref{tab:ablation_cluster} shows, the performance of our clustering procedure is either on par ($D=2$) or superior ($D=3$,$4$) to K-means clustering. In particular, as the number of clusters grows, the performance of K-means drops (\ie from 88.36\% with $D=2$ to 87.21 with $D=4$), while our method - with the same residual GCN - shows performance improvements (\ie from 88.03\% with $D=2$ to 88.74 with $D=4$). This indicates the effectiveness of our local clustering procedure that, differently from K-means, captures the presence of different domains within each client, without requiring one specific model per client.%

Then, we analyze the effect of different initialization strategies for the adjacency matrix of the GCN, considering two choices, \ie domains either disconnected (identity matrix, \textit{eye}) or randomly connected (random adjacency matrix, \textit{rand}). From Table \ref{tab:ablation_cluster}, it is easy to see our method performances are not dependent on the particular initialization strategy, achieving over 88.5\% for all choices with $D=4$. With random initialization though, the performance does not grow with the number of domains, which may indicate the importance of carefully initializing $\mathbf{A}$ as the number of domains grows. For this reason, in the following we always consider a uniform initialization strategy. Note that such a strategy allows the model to refine the domain-specific components separately before merging them based on their distance (see Eq.~\eqref{eq:edge_weight}).

As a third analysis, we focus on the impact of performing a soft combination of domains at test time (as described in Section \ref{sec:specific-models}) rather than using a hard-assignment derived from the predictions of the domain classifier. In both cases, performances are close for all $D$, showing the domain classifier provides reliable domain predictions at test time. In the following, we always consider the soft-assignment due to its higher flexibility. %

Finally, we test the application of the domain-specific modules only on the last layer of the network (rather than on all layers) to see whether a good performance can be achieved while reducing the number of parameters required by \methodShort. %
As the experiments show, using domain-specific parameters on the last layer provides the best results overall (89.18\% with $D=4$), improving the best combination by 0.44\%. Since this choice allows \methodShort\ to use less parameters while still achieving good results%
, we limit the use of domain-specific parameters to the last layers in the next section.

\begin{table}[t]
    \begin{center}
    
    \resizebox{\linewidth}{!}{
    \begin{tabular}{l | c c c |c}
        Model & $\textbf{A}$ & $\textbf{W}$ & ReLU & Acc(\%) \\\hline\hline
        Domain-specific models & - & - & - & 33.61\\\hline
        \multirow{4}{*}{GCN} %
         & U & \ding{55} & \ding{55} & 84.39\\
         & H & \ding{55} & \ding{55} & 82.10\\
         & U & \ding{51} & \ding{55} & 87.92\\
         & H & \ding{51} & \ding{55} & 84.25\\\hline
        \multirow{4}{*}{\methodShort} & U & \ding{51} & \ding{55} & 86.96\\
         & H & \ding{51} & \ding{55} & 88.65\\
         & U & \ding{51} & \ding{51} & 87.97\\
         & H & \ding{51} & \ding{51} & 89.57\\\hline
    \end{tabular}}
    \end{center}
    \vspace{-5pt}
    \caption{\textbf{Ablation studies on CelebA dataset with $N=32$ domains extracted from images meta-data}. $\textbf{A}$ is the adjacency matrix that weights the domains contributions: the symbols (eye,U,H) respectively stand for identity, uniform and weighted (with inverse Hamming distance) matrices. $\textbf{W}$ is the weight projection matrix and ReLU the chosen non-linear activation.}
    \label{tab:ablation_32}
\end{table}

\begin{table}[t]
    \begin{center}
    \resizebox{\linewidth}{!}{
    \begin{tabular}{c |c c c c |c}
        \hline
        \methodShort\ layers & $\textbf{A}$ init & Clusters & $D$ & Soft domains & Acc(\%)\\\hline\hline
        \multirow{12}{*}{all} & eye & K-means & 2 & \ding{55} & 88.36\\
       & eye & K-means & 3 & \ding{55} & 87.97\\
         & eye & K-means & 4 & \ding{55} & 87.21\\\cline{2-6}
         & eye & Clf & 2 & \ding{55} & 88.03\\
         & eye & Clf & 3 & \ding{55} & 88.59\\
         & eye & Clf & 4 & \ding{55} & 88.74\\\cline{2-6}
         & rand & Clf & 2 & \ding{55} & 88.73\\
          & rand & Clf & 3 & \ding{55} & 88.24\\
          & rand & Clf & 4 & \ding{55} & 88.55\\\cline{2-6}
         & eye & Clf & 2 & \ding{51} & 87.88\\
         & eye & Clf & 3 & \ding{51} & 88.74\\
         & eye & Clf & 4 & \ding{51} & 88.67\\\hline
         \multirow{4}{*}{last} & eye & Clf & 2 & \ding{51} & 88.31\\
         & eye & Clf & 3 & \ding{51} & 88.13\\
         & eye & Clf & 4 & \ding{51} & \textbf{89.18}\\
         & eye & Clf & 32 & \ding{51} & 88.40\\\hline
    \end{tabular}}
    \end{center}
    \vspace{-5pt}
    \caption{\textbf{Ablation studies on CelebA dataset with domains given by a priori knowledge or online clustering procedures}. In the $\textbf{A}$ init column, ``eye'' stands for identity matrix and ``rand'' for random. The third column specifies the clustering, \ie clusters generated with K-means or the teacher-student classifier (``Clf'').}
    \vspace{-15pt}
    \label{tab:ablation_cluster}
\end{table}

\subsection{Comparison with the state of the art}
Here we compare our \methodShort\ with state-of-the-art results on both CelebA and FEMNIST. Unfortunately, since different methods employ different settings and client splits, it is difficult to provide an extensive comparison on these datasets. For this reason, on CelebA we compare our model directly with the FedAvg baseline, while for FEMNIST we compare it with FedAvg, FedProx \cite{li2018federated} and SCAFFOLD \cite{karimireddy2020scaffold}. FedProx \cite{li2018federated} adds a proximal term to the standard FedAvg algorithm for improving the model stability when applied over heterogeneous systems and data. SCAFFOLD \cite{karimireddy2020scaffold} uses variance reduction for minimizing the impact of the drift in the updates of each client. We report the results of FedProx and SCAFFOLD shown in their original papers, while for FedAvg we use our baseline. For our method, we use the domain-specific parameters applied on the last layer, $D=4$, soft domain assignments at test time and the adjacency matrix initialized as identity. 

The experimental comparison is reported in Table \ref{tab:exp}. \methodShort\ largely outperforms FedAvg in both scenarios. It achieves $89.18\%$ accuracy compared to 86.88\% of FedAvg on CelebA, and $83.41\%$ accuracy compared to 77.81\% of FedAvg on FEMNIST. This latter improvement (+5.6\%) is remarkable given the higher complexity of the classification task in FEMNIST. Comparing \methodShort\ with FedProx and SCAFFOLD on FEMNIST, we can see that \methodShort\ outperforms FedProx by a large margin (+8.41\%) while being slightly inferior to SCAFFOLD (\ie -0.79\%). However, both FedProx and SCAFFOLD present results under different federated protocols, %
\eg FedProx runs the algorithm for 200 rounds of 10 clients while SCAFFOLD performs 1000 rounds with 20 clients each. %
Despite that, our comparisons demonstrate that \methodShort\ is far superior to the standard FedAvg baselines, due to its better ability to address the statistical heterogeneity across clients, while showing either superior (\wrt FedProx) or competitive (\wrt SCAFFOLD) results with other state-of-the-art algorithms trained on different settings.

As a final analysis, we verify the role of the domain-specific components by checking the final values of the $\lambda$ scalar of Eq.~\eqref{eq:domain-specific-layers}, which weighs the importance of the domain-specific residual. Interestingly, in CelebA, where the concept of heterogeneity is less marked, the $\lambda$ value in the last convolutional layer is 0.3. The final value for FEMNIST, instead, where the heterogeneity across clients is clearer due to the different writing styles, is 2.1. That shows \methodShort\ tailors the use of the domain-specific residual to the specific characteristics of the target dataset and the consequent heterogeneity across the discovered domains. %

\begin{table}[t!]
    \begin{center}
    \begin{tabular}{l l c}
        \hline
        \textbf{Dataset} & \textbf{Model} & \textbf{Accuracy (\%)}\\\hline
        \multirow{2}{*}{CelebA} %
        & FedAvg & 86.88\\
        & \methodShort & \textbf{89.18}\\\hline
        \multirow{4}{*}{FEMNIST} %
        & FedAvg & 77.81\\
        & \textbf{\methodShort} & \textbf{{83.41}}\\\cdashline{2-3} 
        & FedProx & 75.00\\\cdashline{2-3} 
        & SCAFFOLD & 84.20\\\hline
    \end{tabular}
    \end{center}
    \vspace{-2pt}
    \caption{\textbf{Comparison with the state of the art on CelebA and FEMNIST.} %
    We separate the methods according to their setting.%
    }
    \vspace{-15pt}
    \label{tab:exp}
\end{table}
\pdfoutput=1
\section{Conclusions}
In this work, we introduced \methodShort, the first cluster-driven approach addressing statistical heterogeneity in federated learning with Graph Convolutional Neural Networks. \methodShort\ uses an iterative clustering algorithm based on teacher and student domain classifiers. This clustering procedure serves to discover different input distributions, \ie domains, and to instantiate domain-specific parameters accordingly. The domain-specific parameters are connected through a GCN that enables them to interact and share knowledge during training.  These parameters influence the activation of the main, domain-agnostic, network thanks to weighted residual activations. %
Thanks to the domain classifiers and connections of the GCN, new input distributions and unseen users can be addressed at test time via their domain soft-assignment scores. Experimental results show that \methodShort\ outperforms the FedAvg on multiple benchmarks, demonstrating the efficacy of each component. %

{\small 
\textbf{Acknowledgments }
This work has been partially funded by the ERC 853489 - DEXIM, the ERC 802554 - SPECGEO, the ERC 637076 RoboExNovo, the DFG – EXC number 2064/1 – Project number 390727645, and the MIUR under grant ``Dipartimenti di eccellenza 2018-2022''.
}

{\small
\bibliographystyle{ieee_fullname}
\bibliography{egbib}
}

\end{document}